\documentclass[remotesensing,article,submit,pdftex,moreauthors]{Definitions/mdpi}

\firstpage{1}
\pubvolume{1}
\issuenum{1}
\articlenumber{0}
\pubyear{2026}
\copyrightyear{2026}
\datereceived{ }
\daterevised{ }
\dateaccepted{ }
\datepublished{ }


\newcommand{\mbu}{MBU}
\newcommand{\ours}{LER-YOLO}
\newcommand{\rgb}{\mathrm{rgb}}
\newcommand{\ir}{\mathrm{ir}}
\newcommand{\inter}{\mathrm{inter}}
\newcommand{\apfifty}{AP$_{50}$}
\newcommand{\apall}{AP$_{50:95}$}
\newcommand{\yesmark}{\ensuremath{\checkmark}}
\newcommand{\nomark}{\ensuremath{\times}}

\newcommand{\figplaceholder}[1]{\fbox{\parbox[c][0.20\textheight][c]{0.93\linewidth}{\centering\small #1}}}
\graphicspath{{./}{figures/}{remotesensing_figures_ready/}}
\newcommand{\safeincludegraphics}[3][]{%
  \IfFileExists{#2.pdf}{\includegraphics[#1]{#2.pdf}}{%
  \IfFileExists{figures/#2.pdf}{\includegraphics[#1]{figures/#2.pdf}}{%
  \IfFileExists{remotesensing_figures_ready/#2.pdf}{\includegraphics[#1]{remotesensing_figures_ready/#2.pdf}}{%
  \IfFileExists{fig#2.pdf}{\includegraphics[#1]{fig#2.pdf}}{%
  \IfFileExists{figures/fig#2.pdf}{\includegraphics[#1]{figures/fig#2.pdf}}{%
    \figplaceholder{#3. Please provide a vector PDF figure file before submission.}%
  }}}}}%
}

\Title{LER-YOLO: Reliability-Aware Expert Routing for Misaligned RGB-Infrared UAV Detection}

\Author{Liming Hou $^{1}$, Yueping Peng $^{1,*}$, Hexiang Hao $^{1}$, Ji Wang $^{1}$, Xuekai Zhang $^{1}$, Wei Tang $^{1}$, Zecong Ye $^{2}$, Xin Ying $^{1}$ and Yubo He $^{1}$}
\AuthorNames{Liming Hou, Yueping Peng, Hexiang Hao, Ji Wang, Xuekai Zhang, Wei Tang, Zecong Ye, Xin Ying and Yubo He}
\address{$^{1}$ \quad Engineering University of PAP, Xi'an 710086, China; houlimingbeijing@163.com (L.H.); percy001@163.com (Y.P.); hhx1214s@163.com (H.H.); 3202342570@qq.com (J.W.); 18510004596@163.com (X.Z.); 13414667546@163.com (W.T.); 19591587243@163.com (X.Y.); heyubo132@163.com (Y.H.)\\
$^{2}$ \quad Unit Command Department, Officers College of PAP, Chengdu 610213, China; yzc6666@yeah.net (Z.Y.)}
\corres{Correspondence: percy001@163.com}

\makeatletter
\@ifundefined{addhighlights}{%
}{%
  \addhighlights{yes}%
  \renewcommand{\addhighlights}{%
    \vspace{-0.5em}%
    \noindent\textbf{What are the main findings?}\par\vspace{2pt}%
    \begin{itemize}[leftmargin=*,labelsep=2.5mm,topsep=0pt,itemsep=1pt,parsep=0pt]
      \item A Reliability-Guided Sparse Mixture-of-Experts (MoE) Fusion module is proposed to shift RGB-infrared feature fusion from static aggregation to dynamic top-$k$ expert routing.
      \item An Uncertainty-Aware Target Alignment (U-TA) module estimates a spatial reliability map, which serves as an explicit prior for selecting reliable cross-modal interaction and suppressing unreliable fusion.
    \end{itemize}
    \vspace{0pt}%
    \noindent\textbf{What are the implications of the main findings?}\par\vspace{2pt}%
    \begin{itemize}[leftmargin=*,labelsep=2.5mm,topsep=0pt,itemsep=1pt,parsep=0pt]
      \item By using reliability-guided expert routing, \ours{} improves detection accuracy while keeping the parameter count and FLOP budget close to the static-fusion baseline.
      \item Under the reproduced YOLOv5s-family protocol, \ours{} improves the static-fusion baseline from 87.8$\pm$0.2\% to 89.7$\pm$0.2\% \apfifty{}, suggesting that Reliability-Guided Sparse MoE Fusion can be more effective than static fusion for misaligned RGB-infrared UAV detection.
    \end{itemize}%
  }%
}
\makeatother

\abstract{Detecting small unmanned aerial vehicles from RGB infrared remote sensing pairs remains challenging due to tiny target scale, cluttered backgrounds, and spatial misalignment between heterogeneous sensors. Existing bimodal detectors often align or fuse features without assessing the reliability of local cross sensor correspondence, allowing mismatch artifacts to propagate into the detection head. To address this issue, we propose \ours{}, a reliability aware sparse mixture of experts framework for misaligned RGB infrared UAV detection. \ours{} first introduces an Uncertainty Aware Target Alignment module that resamples visible features toward the infrared reference and estimates a spatial reliability map. This reliability prior is then used by a Reliability Guided Sparse MoE Fusion module to adaptively select $k$ experts from RGB dominant, infrared dominant, and interactive fusion experts, enabling trustworthy cross modal interaction while suppressing unreliable fusion. Experiments on the public \mbu{} benchmark under a YOLOv5s family protocol show that \ours{} achieves 89.7$\pm$0.2\% \apfifty{} over three independent seeds, with a best result of 89.9\%. Extensive ablations, parameter matched comparisons, synthetic shift evaluations, and complexity analysis demonstrate that the gains mainly come from reliability guided expert routing rather than increased model capacity.}

\keyword{remote sensing; UAV target detection; RGB-infrared fusion; visible-thermal sensing; misaligned multimodal perception; reliability-aware fusion; sparse mixture-of-experts; expert routing; YOLOv5s; MBU dataset}

\begin{document}

\section{Introduction}

Small unmanned aerial vehicles (UAVs) are increasingly important targets in low altitude remote sensing, with applications in traffic monitoring, infrastructure inspection, disaster response, airspace management, and public safety surveillance. Detecting UAVs from remote sensing imagery is challenging because UAV targets are often tiny, fast moving, weakly textured, and easily confused with cluttered backgrounds such as buildings, trees, clouds, birds, and strong sunlight. Similar difficulties have also been observed in infrared small object detection, where targets usually exhibit low contrast and limited spatial extent~\cite{Dong2025Survey,Ye2026MBUDet,Yang2025IRSmallObject}. Visible images provide rich structural contours and texture details, while infrared images capture thermal responses that remain informative under low illumination or nighttime conditions. Therefore, RGB infrared sensing offers a practical solution for robust UAV detection in complex remote sensing scenarios. This work focuses only on perception level UAV detection using public benchmark data and does not involve interception, engagement, or weaponized counter UAV operations.

Despite its potential, RGB infrared UAV detection remains difficult because practical dual sensor systems are rarely perfectly registered. Sensor baseline differences, lens distortion, platform vibration, gimbal jitter, and asynchronous exposure can all introduce spatial offsets between visible and infrared images. For small UAVs, even a displacement of a few pixels may move the target response into a different local feature region. As a result, direct feature concatenation or addition may create false cross modal correspondence and introduce mismatch artifacts into the detector. The public Misaligned Bimodal UAV (\mbu{}) benchmark explicitly highlights this problem by constructing an RGB infrared detection benchmark from Anti UAV, where target offsets and modality dependent visibility variations are common~\cite{Jiang2021AntiUAV,Ye2026MBUDet}.

Recent multimodal UAV detectors have made notable progress in alignment and fusion. For example, MBUDet integrates target alignment with RGB infrared detection by generating target offset labels~\cite{Ye2026MBUDet}. Other UAV and multispectral remote sensing detectors employ weak alignment, transformer based interaction, dynamic fusion, or lightweight single stage detection to improve accuracy and efficiency~\cite{Chen2024WMFAFA,Liu2025CODAF,Xiao2024GMDETR,Zhao2024GYOLO,Ding2024LRIYOLO}. However, two limitations remain. First, most alignment modules treat the resampled feature as reliable once alignment is completed, although local correspondence can still be uncertain around tiny targets, occlusion boundaries, motion blur, and background distractors. Second, many fusion modules rely on a fixed operator or static attention rule, which is insufficient because the reliability of RGB and infrared cues varies across scenes and spatial locations. These observations suggest that effective bimodal UAV detection should not only align features, but also estimate whether the aligned correspondence is trustworthy before fusion.

To address these issues, we propose \ours{} (Local Reliability Expert Routing YOLO), a reliability aware RGB infrared UAV detection framework for misaligned remote sensing imagery. The core idea is to make local alignment reliability an explicit routing prior for multimodal fusion. Specifically, we introduce an Uncertainty Aware Target Alignment (U TA) module that resamples visible features toward the infrared reference and simultaneously predicts a spatial reliability map. This map estimates whether the aligned RGB feature is trustworthy at each spatial location. Based on this reliability prior, we further design a Reliability Guided Sparse MoE Fusion module, which adaptively selects top $k$ experts from three heterogeneous candidates, including an RGB dominant expert, an infrared dominant expert, and an interactive fusion expert. In this way, reliable regions can exploit cross modal interaction, while unreliable regions can rely more on modality specific evidence. Our goal is not to redesign the detection backbone, but to improve the decision mechanism of RGB infrared interaction after alignment while avoiding unnecessary dense expert execution.

The main contributions are summarized as follows:
\begin{itemize}
\item We formulate misaligned RGB infrared UAV detection as a reliability aware fusion problem, where local alignment trustworthiness explicitly controls downstream cross modal interaction.

\item We propose an Uncertainty Aware Target Alignment (U TA) module that jointly performs visible to infrared feature resampling and spatial reliability estimation, enabling unreliable aligned features to be identified before fusion.

\item We design a Reliability Guided Sparse MoE Fusion module that adaptively routes features to modality specific and interactive experts, shifting RGB infrared fusion from static aggregation to reliability guided expert selection.

\item We conduct a comprehensive evaluation on the public \mbu{} protocol, including seed level analysis, ablation studies, parameter matched comparison, controlled synthetic misalignment, model complexity reporting, and supplementary log and statistics provenance.
\end{itemize}

The rest of this paper is organized as follows. Section~\ref{sec:related} reviews UAV detection, RGB infrared fusion, and reliability aware routing. Section~\ref{sec:method} presents the proposed framework. Section~\ref{sec:experiments} reports the experiments and analysis. Section~\ref{sec:conclusion} concludes the paper.

\section{Related Work}
\label{sec:related}

\subsection{UAV and Anti UAV Target Detection}

Deep object detectors have been widely adopted for UAV remote sensing because they can localize small targets efficiently in complex aerial scenes. Among them, the YOLO family is particularly suitable for deployment oriented perception, since one stage detection avoids proposal generation and offers a favorable balance between accuracy and efficiency~\cite{Redmon2016YOLO,Dadboud2021YOLOv5,Tang2024Survey,Dai2021Overview}. Recent studies on UAV and infrared remote sensing further improve lightweight detection, small target representation, and real time inference, including ITD YOLOv8, G YOLO, LRI YOLO, all time UAV detection, high altitude infrared detection, and multi scale thermal feature extraction~\cite{Zhao2024ITDYolo,Zhao2024GYOLO,Ding2024LRIYOLO,Huang2024AllTime,Huang2025RTDETR,Wang2025MDDFA}. Meanwhile, newly released UAV benchmarks have expanded the evaluation scope from single modality detection to registration, tracking, condition diversity, and RGB thermal aerial perception~\cite{Bin2025ATRUMMIM,Chen2025ConditionCues,Qin2025MUST,Lee2024CaltechAerial,Xie2025CSTAntiUAV,Gross2026SegFly}. These efforts show that practical UAV detection must jointly consider detection accuracy, model complexity, and robustness under diverse sensing conditions. Anti UAV detection is more challenging because targets are usually tiny, weakly textured, fast moving, and easily confused with birds or background structures~\cite{Lyu2023TinyAirborne,Munir2024RainyArtifacts,Dong2025Survey}.

\subsection{RGB Infrared and Visible Thermal Fusion}

RGB infrared fusion combines visible appearance with infrared thermal responses, providing complementary cues for remote sensing perception under low illumination, thermal crossover, clutter, and long range observation~\cite{Pereira2024IRVisible,Svanstrom2022DroneFusion,ElAhmar2023EnhancedFusion}. Existing visible thermal detectors and fusion frameworks have explored transformer interaction, prompt tuning, attention based fusion, Mamba style modeling, image registration, semantic fusion, and multimodal benchmark construction~\cite{Chen2024VIPDet,Park2024CMTFusion,Wang2024CrossModalAttention,Zhu2024TaskMoA,Ying2025VTTiny,Li2025DEPFusion,Liu2024COMO,Nguyen2025CfYOLO,Zuo2025SFFR,Li2024NoStrictRegistration,Tang2022SuperFusion,Xie2023SemLA,Guo2025DPDETR}. In a broader vision context, conditional modeling has also shown strong potential for preserving controllability and consistency across heterogeneous visual conditions, such as pose guided person generation, customizable virtual dressing, long term talking face generation, and story visualization~\cite{shen2024imagpose,shen2024advancing,shen2025imagdressing,shenlong,shen2025boosting}. These works suggest that explicitly modeling condition dependent information is important for robust visual synthesis and perception. However, most RGB infrared detectors still assume that the two modalities are aligned or only weakly misaligned. This assumption becomes fragile in anti UAV scenarios, where even a few pixels of displacement may separate the visible and infrared responses of a tiny target.

\subsection{Misaligned Multimodal Detection and Alignment}

Spatial misalignment has motivated a series of feature alignment techniques, including deformable convolution, spatial transformer networks, offset prediction, correlation modeling, calibrated transformers, and cross modal calibration~\cite{Dai2017DCN,Jaderberg2015STN,He2023ADCNet,Tu2022DCNet,Song2023MROS,Yuan2024C2Former,Yuan2024CascadeAlignment,Yuan2022TSRAlignment}. For UAV detection, MBUDet constructs a dedicated misaligned bimodal benchmark and introduces target offset label generation for supervised target level alignment~\cite{Ye2026MBUDet}. Weakly alignment free adaptive feature alignment and cross modal offset guided dynamic alignment further demonstrate the importance of explicit spatial reasoning in UAV based multimodal detection~\cite{Chen2024WMFAFA,Liu2025CODAF}. Nevertheless, most existing alignment based methods treat the aligned feature as reliable after resampling. In practice, the estimated correspondence may still be uncertain around tiny targets, occlusions, motion blur, depth discontinuities, and background distractors. Therefore, alignment alone is insufficient. A detector also needs to estimate whether the aligned local correspondence should be trusted before deep fusion.

\subsection{Reliability Estimation and Sparse MoE Fusion}

Uncertainty estimation provides a principled way to measure whether a local prediction or representation is trustworthy~\cite{Kendall2017Uncertainty,Cho2024Cocoon}. This is especially important for multimodal perception, since different sensors may fail under different illumination, thermal contrast, weather, and motion conditions. Mixture of Experts (MoE) models offer a flexible mechanism for conditional computation by assigning different inputs or spatial regions to specialized experts~\cite{Shazeer2017MoE,Chen2023AdaMVMoE,Zhu2024TaskMoA}. Recent MoE based perception methods extend expert routing to adaptive object detection, multimodal 3D understanding, and all weather multimodal perception~\cite{Meiraz2025YOLOMoE,Li2025MoE3D,Lin2026AWMoE}. However, many existing routers rely mainly on semantic features or task level cues, without explicitly considering whether local cross modal correspondence is reliable. In contrast, our method uses alignment reliability as a spatial routing prior. It encourages cross modal interaction in trustworthy regions while suppressing unreliable fusion where misalignment artifacts may harm UAV localization.

\section{Method}
\label{sec:method}

\subsection{Problem Formulation and Overall Framework}

This section describes the materials and methods used for the proposed remote-sensing detector. Let $I_{\rgb}\in\mathbb{R}^{3\times H_0\times W_0}$ and $I_{\ir}\in\mathbb{R}^{3\times H_0\times W_0}$ denote an unregistered visible image and an infrared image, respectively. The goal is to detect UAV targets from the misaligned RGB-infrared pair while preserving a practical model size and detector structure. Following MBUDet, the infrared branch is used as the spatial reference because UAV targets are usually more stable in the infrared modality under low illumination \cite{Ye2026MBUDet}. Two non-shared YOLOv5s-family branches extract multi-scale features $F_{\rgb}$ and $F_{\ir}$. The visible feature is aligned to the infrared reference by U-TA, and the aligned visible feature, the infrared feature, and the reliability map are fused by the Reliability-Guided Sparse MoE Fusion module.

\begin{figure}[t]
\centering
\safeincludegraphics[width=0.96\textwidth]{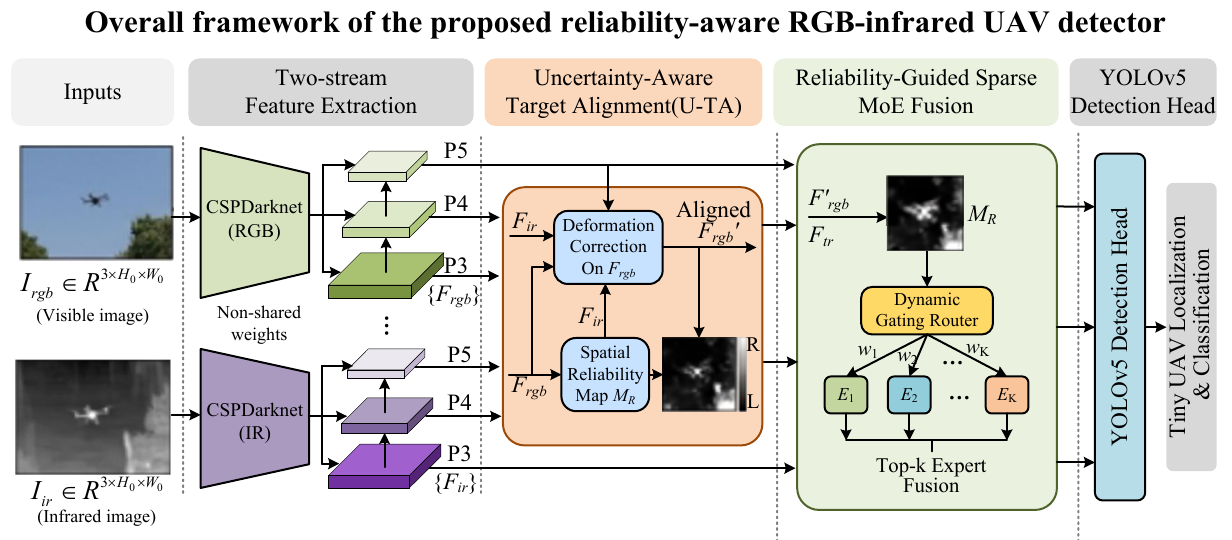}{Final Figure 1: overall architecture}
\caption{Overall pipeline of \ours{}. The framework connects Uncertainty-Aware Target Alignment (U-TA) with Reliability-Guided Sparse MoE Fusion for misaligned RGB-infrared remote-sensing UAV detection. The diagram shows the input pair, U-TA, top-$k$ expert fusion, and the YOLOv5 detection head used in the main experiments.\label{fig:framework}}
\end{figure}

Figure~\ref{fig:framework} summarizes the pipeline. The important design principle is that alignment quality is not discarded after resampling. Instead, it becomes an explicit routing prior that controls how strongly each local remote-sensing region should use modality-specific or cross-modal processing.

\subsection{Uncertainty-Aware Target Alignment (U-TA)}

Given a feature pair $(F_{\rgb},F_{\ir})$, an offset predictor estimates a dense two-dimensional deformation field $\Delta p$. For a spatial position $p_0$ on the infrared reference grid, the aligned visible feature is computed by bilinear resampling:
\begin{equation}
\widetilde{F}_{\rgb}(p_0)=\sum_{k=1}^{K}w_kF_{\rgb}(p_0+p_k+\Delta p_k)\,,
\label{eq:resampling}
\end{equation}
where $\{p_k\}_{k=1}^{K}$ is the regular sampling grid, $w_k$ is the sampling weight, and $\Delta p_k$ is the learned offset. This operation pulls displaced visible responses toward the infrared target coordinate. The supervision design follows the same spirit as target-center and template-based alignment strategies used in object detection and registration, where local target response maps can provide a more stable learning signal than raw image-level registration \cite{Law2018CornerNet,Duan2019CenterNet,Hashemi2016TemplateMatching}.

Local resampling reduces spatial discrepancy but does not guarantee reliable correspondence. Therefore, a parallel reliability branch predicts a continuous reliability map:
\begin{equation}
R=\sigma\left(\phi\left([F_{\ir},\widetilde{F}_{\rgb}]\right)\right)\,,
\label{eq:reliability}
\end{equation}
where $\phi(\cdot)$ denotes the reliability subnetwork, $[\cdot,\cdot]$ denotes channel concatenation, and $\sigma(\cdot)$ is the sigmoid function. A high value of $R(x,y)$ indicates trustworthy local correspondence after alignment, whereas a low value indicates mismatch, occlusion, or unstable resampling.

\begin{figure}[t]
\centering
\safeincludegraphics[width=0.90\textwidth]{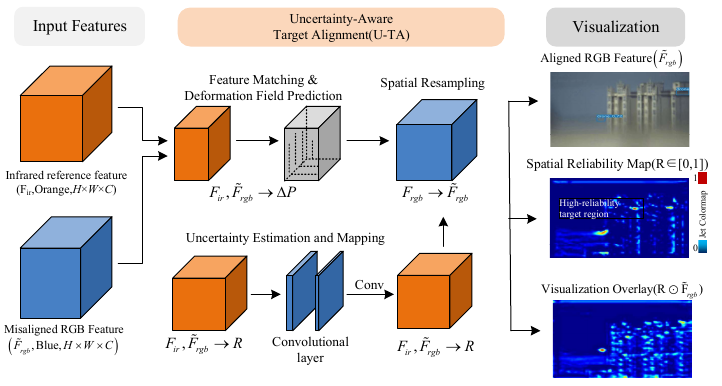}{Final Figure 2: U-TA module}
\caption{Uncertainty-Aware Target Alignment (U-TA). The module aligns visible features to the infrared reference and predicts a spatial reliability map, where higher reliability indicates more trustworthy local RGB-infrared correspondence.\label{fig:uta}}
\end{figure}

To supervise reliability estimation without pixel-level alignment annotations, a self-supervised U-TA reliability loss is used:
\begin{equation}
\mathcal{L}_{uta}=\frac{1}{N}\sum_{i,j}\left(R_{ij}\left\|F^{ij}_{\ir}-\widetilde{F}^{ij}_{\rgb}\right\|_1-\lambda\log(R_{ij}+\epsilon)\right)\,,
\label{eq:uta_loss}
\end{equation}
where $N$ is the number of spatial locations, $\lambda$ is a balancing coefficient, and $\epsilon$ is a small constant. The first term encourages high reliability when the aligned feature is close to the infrared reference, and the logarithmic term avoids the trivial solution of assigning zero reliability everywhere.

\subsection{Reliability-Guided Sparse MoE Fusion}

A single static fusion operator is unlikely to be optimal for all regions. Reliable regions can benefit from cross-modal interaction, whereas unreliable regions should rely more on modality-specific evidence. The proposed Reliability-Guided Sparse MoE Fusion module uses three heterogeneous expert candidates: an RGB-dominant expert $E_{\rgb}$, an infrared-dominant expert $E_{\ir}$, and an interactive fusion expert $E_{\inter}$.

Before routing, the aligned visible feature is modulated by the reliability map:
\begin{equation}
F_{in}=\mathcal{C}\left(F_{\ir},\widetilde{F}_{\rgb}\odot R\right)\,,
\label{eq:router_input}
\end{equation}
where $\mathcal{C}(\cdot)$ is channel concatenation and $\odot$ denotes element-wise multiplication. The router first predicts expert probabilities:
\begin{equation}
g=\mathrm{softmax}(G(F_{in}))\,, \qquad g=\{g_{\rgb},g_{\ir},g_{\inter}\}\,.
\label{eq:expert_weights}
\end{equation}
Then the top-$k$ experts are selected according to the router probabilities:
\begin{equation}
S=\mathrm{TopK}(g,k)\,.
\label{eq:topk_selection}
\end{equation}
The selected experts are executed, and their routing probabilities are re-normalized before aggregation:
\begin{equation}
\widehat{g}_{i}=\frac{g_i}{\sum_{j\in S}g_j},\quad i\in S\,.
\label{eq:renorm_weights}
\end{equation}
The final fused feature is obtained by sparse reliability-weighted expert aggregation:
\begin{equation}
F_{out}=\sum_{i\in S}\widehat{g}_{i}E_i(F_{in})\,.
\label{eq:fusion_output}
\end{equation}

\begin{figure}[t]
\centering
\safeincludegraphics[width=0.88\textwidth]{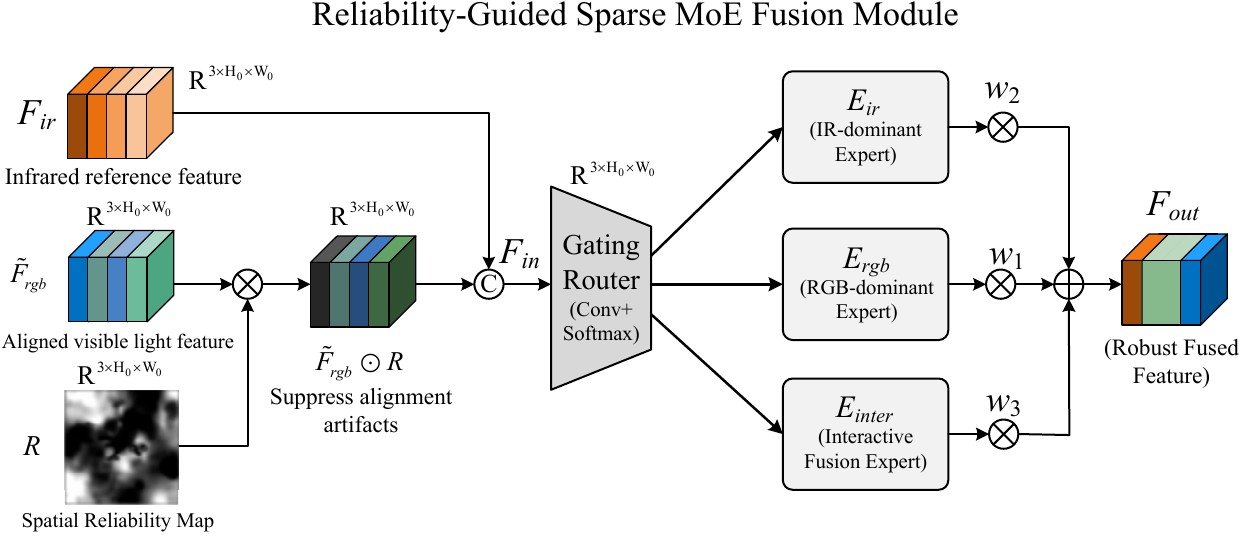}{Final Figure 3: reliability-guided sparse MoE fusion}
\caption{Reliability-Guided Sparse MoE Fusion. The gating router uses the reliability prior to select top-$k$ experts from RGB-dominant, infrared-dominant, and interactive fusion candidates for remote-sensing feature fusion.\label{fig:moe}}
\end{figure}

In the main implementation, the router predicts content-adaptive weights over the expert pool and aggregates the selected top-$k$ experts during training and inference. With the current default setting of $k=3$ and three available experts, all experts are active, but their contributions remain conditionally weighted rather than fixed. This design preserves an extensible sparse-selection mechanism for future larger expert pools while already enabling scene-dependent expert emphasis in the present model.

\subsection{Optimization}

The detector is optimized end-to-end with detection, target-alignment, and U-TA reliability losses:
\begin{equation}
\mathcal{L}=\mathcal{L}_{det}+\alpha\mathcal{L}_{ta}+\beta\mathcal{L}_{uta}\,,
\label{eq:total_loss}
\end{equation}
where $\mathcal{L}_{det}$ is the standard YOLO detection loss, $\mathcal{L}_{ta}$ denotes target-alignment supervision derived from target-offset pseudo labels, and $\mathcal{L}_{uta}$ is the reliability-aware alignment loss. The coefficients $\alpha$ and $\beta$ balance detection and alignment-related objectives.

\subsection{Dataset and Implementation Details}

Experiments are conducted on the public \mbu{} dataset following the MBUDet protocol \cite{Ye2026MBUDet}. The \mbu{} benchmark is derived from the public Anti-UAV bimodal tracking dataset \cite{Jiang2021AntiUAV} by extracting one frame every 40 frames and removing watermark regions. Anti-UAV contains 318 RGB-infrared video pairs, with 160 pairs for training, 67 pairs for validation, and 91 pairs for testing. The resulting \mbu{} benchmark contains 14,946 images: 7530 training images with 7309 targets, 3122 validation images with 2983 targets, and 4294 test images with 4127 targets. The same split and preprocessing protocol are used to ensure direct comparison with MBUDet. This study does not create a new dataset; all dataset images are from public benchmarks, and the generated seed-level summaries, routing-statistics summaries, and synthetic-shift evaluation records are prepared as supplementary material.

All main experiments use YOLOv5s as the base detector, matching the detector family used by MBUDet. The input image size is 320, the batch size is 16, and the main training schedule is 80 epochs. Unless otherwise specified, all main Reliability-Guided Sparse MoE Fusion experiments use $k=3$ over the current three-expert pool, and top-$k$ routing is applied during both training and inference. In the present implementation, $k=3$ activates all currently available experts, but the fusion remains routing-based rather than static because the router still generates input-dependent normalized weights. More importantly, the top-$k$ formulation defines an extensible expert pool: if additional experts such as radar-specific, weather-specific, or platform-specific branches are introduced in future work, the same $k=3$ policy naturally becomes a sparse selective activation strategy. Precision (P), recall (R), F1-score, \apfifty{}, \apall{}, parameter count and FLOPs are reported when available. Since MBUDet and many public baselines mainly report \apfifty{}, \apfifty{} is used as the primary comparison metric. Three independent training seeds of \ours{} produce test-set \apfifty{} values of 0.895, 0.898, and 0.899. The main tables use mean$\pm$standard deviation across independent seeds when the corresponding logs are available; for \ours{}, the three seed-level \apfifty{} values yield 89.7$\pm$0.2\%, with the best run reaching 89.9\%. Seed-level summaries, routing-statistics summaries, and synthetic-shift details are prepared as supplementary material to make the reported results auditable.

\section{Experiments and Analysis}
\label{sec:experiments}

\subsection{Main Comparison on the MBU Benchmark}

For context, the original MBUDet study reports a two-stage training recipe with visible-branch load-and-freeze pretraining, obtaining 90.1\% \apfifty{} on the \mbu{} benchmark \cite{Ye2026MBUDet}. It also compares with representative fusion and detection methods, including SuperYOLO, DEYOLO, CDC-YOLOFusion, CFT, ICAFusion, ProbEn3, and FBRT-YOLO \cite{Zhang2023SuperYOLO,Chen2024DEYOLO,Wang2024CDCYoloFusion,Qingyun2021CFT,Shen2024ICAFusion,Chen2022ProbEn3,Xiao2025FBRTYOLO}. Because these published results may use different training recipes, reporting conventions, or profiling environments, they are treated only as background context rather than included in the controlled comparison table. The main evidence in this paper is the reproduced comparison under the same YOLOv5s-family setting.

\subsection{Controlled Comparison under the Reproduced Protocol}

Table~\ref{tab:main} compares the proposed method with controlled reproduced baselines under the YOLOv5s-family setting. The corrected U-TA static-fusion baseline obtains 87.8$\pm$0.2\% \apfifty{}, while \ours{} obtains 89.7$\pm$0.2\% \apfifty{}. Therefore, within the reproduced protocol, the improvement is evaluated against the controlled static-fusion baseline rather than presented as a universal best-performing result.

\begin{table}[t]
\caption{Controlled reproduced comparison on the \mbu{} dataset under the YOLOv5s-family setting. Runtime metrics are omitted because speed is not used as an advantage claim.\label{tab:main}}
\centering
\scriptsize
\setlength{\tabcolsep}{2.0pt}
\renewcommand{\arraystretch}{1.13}
\begin{tabularx}{\textwidth}{@{}>{\raggedright\arraybackslash}p{0.165\textwidth}>{\centering\arraybackslash}p{0.075\textwidth}*{6}{>{\centering\arraybackslash}X}@{}}
\toprule
\textbf{Method} & \textbf{Modality} & \textbf{P (\%)} & \textbf{R (\%)} & \textbf{F1 (\%)} & \textbf{\apfifty{} (\%)} & \textbf{\apall{} (\%)} & \textbf{Params (M)} \\
\midrule
YOLOv5s-RGB~\cite{Dadboud2021YOLOv5} & RGB & $90.6\pm1.5$ & $76.1\pm1.6$ & $82.7\pm1.2$ & $80.7\pm0.9$ & $35.0\pm0.8$ & 7.1 \\
YOLOv5s-IR~\cite{Dadboud2021YOLOv5} & IR & $91.1\pm1.5$ & $76.5\pm1.6$ & $83.2\pm1.2$ & $81.2\pm0.7$ & $35.2\pm0.8$ & 7.1 \\
Early-Concat~\cite{Ye2026MBUDet} & RGB+IR & $95.5\pm1.3$ & $82.1\pm2.5$ & $88.3\pm2.0$ & $86.7\pm1.6$ & $41.1\pm0.4$ & 13.6 \\
Channel Add.~\cite{Ye2026MBUDet} & RGB+IR & $94.7\pm1.3$ & $82.7\pm2.0$ & $88.3\pm0.7$ & $86.8\pm0.8$ & $41.2\pm1.0$ & 12.9 \\
U-TA Static & RGB+IR & $95.5\pm1.5$ & $83.6\pm1.3$ & $89.1\pm0.2$ & $87.8\pm0.2$ & $42.1\pm0.5$ & 12.205 \\
PM Static & RGB+IR & $95.4\pm1.1$ & $83.0\pm1.0$ & $88.7\pm0.5$ & $87.5\pm0.7$ & $42.2\pm0.6$ & 12.7 \\
\ours{} & RGB+IR & $96.9\pm0.5$ & $86.5\pm0.5$ & $91.4\pm0.3$ & \textbf{$89.7\pm0.2$} & \textbf{$43.5\pm1.0$} & 12.21 \\
\bottomrule
\end{tabularx}
\vspace{2pt}
\begin{minipage}{\textwidth}
\footnotesize \textbf{Note:} Metrics are reported as mean$\pm$standard deviation over three reproduced runs; Channel Add., U-TA Static, and PM Static denote channel addition, U-TA with static fusion, and parameter-matched static fusion, respectively.
\end{minipage}
\end{table}

\subsection{Component Ablation}

Table~\ref{tab:ablation} evaluates the contribution of the sparse MoE structure and reliability guidance using clearer variant names. The static-fusion baseline uses U-TA but does not use expert routing. Unguided Sparse MoE Fusion removes reliability-guided routing, and Reliability-Guided Static Fusion keeps the reliability cue but removes MoE specialization. These comparisons are intended to isolate the roles of expert specialization and reliability guidance without overstating the strength of any single component.

\begin{table}[t]
\caption{Component ablation under the YOLOv5s-family setting.\label{tab:ablation}}
\centering
\scriptsize
\setlength{\tabcolsep}{1.8pt}
\renewcommand{\arraystretch}{1.13}
\begin{tabularx}{\textwidth}{@{}>{\raggedright\arraybackslash}p{0.165\textwidth}*{7}{>{\centering\arraybackslash}X}@{}}
\toprule
\textbf{Variant} & \textbf{RGB} & \textbf{IR} & \textbf{U-TA} & \textbf{Static fusion} & \textbf{Sparse MoE} & \textbf{RG} & \textbf{\apfifty{} (\%)} \\
\midrule
YOLOv5s-RGB~\cite{Dadboud2021YOLOv5} & \yesmark & \nomark & \nomark & \nomark & \nomark & \nomark & $80.7\pm0.9$ \\
YOLOv5s-IR~\cite{Dadboud2021YOLOv5} & \nomark & \yesmark & \nomark & \nomark & \nomark & \nomark & $81.2\pm0.7$ \\
Early-Concat~\cite{Ye2026MBUDet} & \yesmark & \yesmark & \nomark & \nomark & \nomark & \nomark & $86.7\pm1.6$ \\
U-TA Static & \yesmark & \yesmark & \yesmark & \yesmark & \nomark & \nomark & $87.8\pm0.2$ \\
Unguided MoE & \yesmark & \yesmark & \yesmark & \nomark & \yesmark & \nomark & $89.2\pm0.6$ \\
RG Static & \yesmark & \yesmark & \yesmark & \yesmark & \nomark & \yesmark & $87.2\pm0.5$ \\
PM Static & \yesmark & \yesmark & \yesmark & \yesmark & \nomark & \nomark & $87.5\pm0.7$ \\
\ours{} & \yesmark & \yesmark & \yesmark & \nomark & \yesmark & \yesmark & \textbf{$89.7\pm0.2$} \\
\bottomrule
\end{tabularx}
\vspace{2pt}
\begin{minipage}{\textwidth}
\footnotesize \textbf{Note:} U-TA and RG denote Uncertainty-Aware Target Alignment and reliability guidance, respectively; U-TA Static, Unguided MoE, RG Static, and PM Static denote the corresponding ablation variants.
\end{minipage}
\end{table}

\subsection{Top-$k$ Sensitivity of Expert Routing}

Because the number of activated experts directly affects routing behavior, we evaluate $k\in\{1,2,3\}$ in Table~\ref{tab:topk_sensitivity}. The three settings produce comparable results, indicating that the proposed routing mechanism is not highly sensitive to $k$. Among them, $k=3$ gives the highest mean \apfifty{} and AP$_{75}$, although the margin over $k=1$ is small. Since \apfifty{} is the primary metric used for the \mbu{} benchmark comparison, the final reported model adopts $k=3$ as the default configuration. This experiment is therefore used for sensitivity checking and configuration selection, not as a strong statistical superiority claim. In the current three-expert implementation, $k=3$ activates the full expert set, but the fusion is still routing-based rather than static because the RGB-dominant, infrared-dominant, and interactive experts receive input-dependent weights. When future experts such as radar-specific, weather-specific, or platform-specific branches are added, keeping $k=3$ will naturally become selective sparse activation over a larger candidate pool while preserving the same dynamic-routing rule.

\begin{table}[t]
\caption{Sensitivity analysis of the number of active experts $k$ in Reliability-Guided Sparse MoE Fusion.}\label{tab:topk_sensitivity}
\centering
\footnotesize
\setlength{\tabcolsep}{4.0pt}
\renewcommand{\arraystretch}{1.12}
\begin{tabularx}{\textwidth}{@{}>{\centering\arraybackslash}p{0.10\textwidth}*{4}{>{\centering\arraybackslash}X}@{}}
\toprule
\textbf{$k$} & \textbf{\apfifty{} (\%)} & \textbf{AP$_{75}$ (\%)} & \textbf{\apall{} (\%)} & \textbf{Active Params (M)} \\
\midrule
1 & $87.59\pm1.38$ & $30.34\pm3.00$ & $41.93\pm1.44$ & 12.2059 \\
2 & $87.32\pm0.13$ & $29.85\pm2.50$ & $41.44\pm1.17$ & 12.2061 \\
3 & \textbf{$87.62\pm1.13$} & $30.43\pm1.63$ & $41.70\pm0.84$ & 12.2063 \\
\bottomrule
\end{tabularx}
\vspace{2pt}
\begin{minipage}{\textwidth}
\footnotesize \textbf{Note:} Mean \apfifty{} is highlighted; close results indicate low sensitivity to $k$.
\end{minipage}
\end{table}

\subsection{Parameter-Matched Static-Fusion Comparison}

A potential concern for MoE-style fusion is that improvement may come from increased capacity. To address this, Table~\ref{tab:parammatched} uses a Parameter-Matched Static Fusion baseline, which widens the static fusion branch to approach the parameter count of \ours{} while removing sparse expert routing. This comparison is kept separate from the main comparison because it is a capacity-control variant rather than a standard baseline.

\begin{table}[t]
\caption{Parameter-matched static-fusion comparison under the same test protocol.\label{tab:parammatched}}
\centering
\footnotesize
\setlength{\tabcolsep}{4.0pt}
\renewcommand{\arraystretch}{1.12}
\begin{tabularx}{\textwidth}{@{}>{\raggedright\arraybackslash}p{0.34\textwidth}*{4}{>{\centering\arraybackslash}X}@{}}
\toprule
\textbf{Method} & \textbf{Params (M)} & \textbf{FLOPs (G)} & \textbf{\apfifty{} (\%)} & \textbf{\apall{} (\%)} \\
\midrule
U-TA + Static Fusion & 12.205 & 7.4 & $87.8\pm0.2$ & $42.1\pm0.5$ \\
Parameter-Matched Static Fusion & 12.7 & 8.1 & $87.5\pm0.7$ & $42.2\pm0.6$ \\
\ours{} & 12.21 & 7.5 & \textbf{$89.7\pm0.2$} & \textbf{$43.5\pm1.0$} \\
\bottomrule
\end{tabularx}
\vspace{2pt}
\begin{minipage}{\textwidth}
\footnotesize \textbf{Note:} This capacity-control baseline removes sparse expert routing.
\end{minipage}
\end{table}

\subsection{Hyperparameter Sensitivity}

Table~\ref{tab:hyperparam} reports the sensitivity analysis of the alignment and reliability-related loss weights under the 80-epoch hyperparameter-sweep protocol. The sweep changes one coefficient at a time around the default sweep setting while keeping the remaining coefficients fixed. Across the tested range, \apfifty{} varies from 87.1\% to 88.1\%, indicating that moderate changes in $\alpha$, $\beta$, and $\lambda$ do not lead to large performance fluctuations. A smaller $\beta$ slightly improves \apfifty{}, whereas the default setting achieves the highest \apall{} within this sweep. Therefore, the default configuration is retained as a balanced setting rather than claimed to be the unique optimum.

\begin{table}[t]
\caption{Sensitivity analysis of loss-balancing hyperparameters under the 80-epoch hyperparameter-sweep protocol.\label{tab:hyperparam}}
\centering
\footnotesize
\setlength{\tabcolsep}{3.0pt}
\renewcommand{\arraystretch}{1.13}
\begin{tabularx}{\textwidth}{@{}>{\centering\arraybackslash}p{0.13\textwidth}*{5}{>{\centering\arraybackslash}X}@{}}
\toprule
\textbf{Sweep} & \textbf{$\alpha$} & \textbf{$\beta$} & \textbf{$\lambda$} & \textbf{\apfifty{} (\%)} & \textbf{\apall{} (\%)} \\
\midrule
$\alpha$ & $0.5$ & $1.0$ & $1.0{\times}10^{-4}$ & $87.6\pm0.3$ & $41.7\pm1.2$ \\
Default\textsuperscript{\dag} & $1.0$ & $1.0$ & $1.0{\times}10^{-4}$ & $87.8\pm0.6$ & \textbf{$43.0\pm1.1$} \\
$\alpha$ & $1.5$ & $1.0$ & $1.0{\times}10^{-4}$ & $87.2\pm0.8$ & $41.9\pm2.2$ \\
\midrule
$\beta$ & $1.0$ & $0.5$ & $1.0{\times}10^{-4}$ & \textbf{$88.1\pm0.3$} & $42.2\pm1.4$ \\
$\beta$ & $1.0$ & $1.5$ & $1.0{\times}10^{-4}$ & $87.2\pm0.7$ & $41.2\pm0.8$ \\
\midrule
$\lambda$ & $1.0$ & $1.0$ & $5.0{\times}10^{-5}$ & $87.4\pm0.6$ & $42.2\pm0.4$ \\
$\lambda$ & $1.0$ & $1.0$ & $1.5{\times}10^{-4}$ & $87.1\pm0.5$ & $41.3\pm1.8$ \\
\bottomrule
\end{tabularx}
\vspace{2pt}
\begin{minipage}{\textwidth}
\footnotesize \textbf{Note:} One factor is varied at a time; the default row is shared across sweeps.
\end{minipage}
\end{table}

\subsection{Robustness to Synthetic Cross-Modal Misalignment}

To evaluate spatial-mismatch robustness, the RGB branch is shifted during testing while the infrared branch and labels are unchanged. This perturbation simulates sensor-baseline variation and platform jitter. For each nonzero shift magnitude $s$, the four directions $(s,0)$, $(-s,0)$, $(0,s)$, and $(0,-s)$ are averaged. Table~\ref{tab:shift} reports the aligned condition using the same three-seed summary as Table~\ref{tab:main}, and the shifted conditions using the available repeated evaluations for the U-TA static-fusion baseline and \ours{}. Under the aligned condition, \ours{} improves both \apfifty{} and \apall{}. Under synthetic shifts, \ours{} maintains positive \apall{} gains, while \apfifty{} becomes comparable to the static baseline as the shift increases. We therefore interpret the robustness results conservatively: reliability-aware routing provides a clear aligned-condition gain and maintains positive \apall{} gains under synthetic shifts, while \apfifty{} becomes comparable to the static baseline as the shift increases.

\begin{table}[t]
\caption{Robustness under synthetic RGB-to-IR misalignment. The 0 px row uses the aligned-test summary for \ours{}; nonzero-shift rows are averaged over available shift directions at each magnitude.\label{tab:shift}}
\centering
\scriptsize
\setlength{\tabcolsep}{2.8pt}
\renewcommand{\arraystretch}{1.14}
\begin{tabularx}{\textwidth}{@{}>{\centering\arraybackslash}p{0.09\textwidth}*{6}{>{\centering\arraybackslash}X}@{}}
\toprule
\textbf{Shift} & \shortstack{\textbf{U-TA + Static}\\\textbf{\apfifty{}}} & \shortstack{\textbf{Ours}\\\textbf{\apfifty{}}} & \textbf{Gain} & \shortstack{\textbf{U-TA + Static}\\\textbf{\apall{}}} & \shortstack{\textbf{Ours}\\\textbf{\apall{}}} & \textbf{Gain} \\
\midrule
0 px  & $87.8\pm0.2$ & $89.7\pm0.2$ & +1.9 & $42.1\pm0.5$ & $43.5\pm1.0$ & +1.4 \\
5 px  & $87.3\pm0.2$ & $87.3\pm0.3$ & $-0.1$ & $41.3\pm0.3$ & $41.9\pm0.4$ & +0.6 \\
10 px & $87.2\pm0.6$ & $87.0\pm0.4$ & $-0.2$ & $41.1\pm0.3$ & $41.9\pm0.4$ & +0.7 \\
20 px & $86.4\pm0.5$ & $86.2\pm0.6$ & $-0.2$ & $41.1\pm0.4$ & $41.5\pm0.7$ & +0.4 \\
40 px & $85.8\pm0.5$ & $85.8\pm0.4$ & +0.0 & $40.6\pm1.0$ & $40.9\pm0.8$ & +0.3 \\
\bottomrule
\end{tabularx}
\end{table}

\subsection{Qualitative and Interpretability Analysis}

\begin{figure}[t]
\centering
\safeincludegraphics[width=0.90\textwidth]{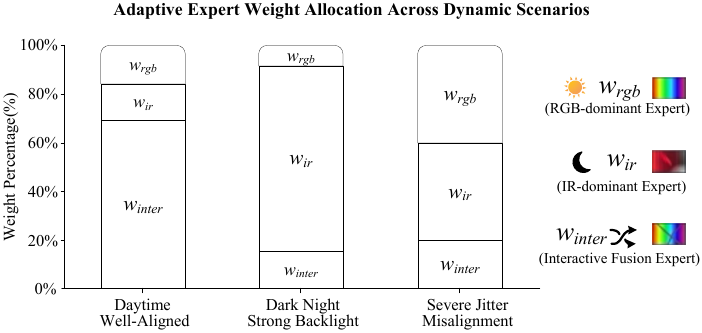}{Final Figure 4: conceptual expert-weight allocation}
\caption{Conceptual illustration of adaptive expert-weight allocation under representative UAV remote-sensing scenarios. The diagram summarizes the intended behavior of reliability-guided routing: stronger interactive fusion in reliable daytime scenes, stronger infrared-dominant processing under dark or backlit conditions, and reduced risky interaction under severe jitter or misalignment.\label{fig:routing}}
\end{figure}

Figure~\ref{fig:routing} illustrates the expected expert-allocation tendency under representative degradation conditions. This figure is used as a conceptual illustration of the routing mechanism rather than as a statistical claim: in well-aligned daytime scenes, the interaction expert is expected to contribute more; in dark or strongly backlit scenes, the infrared-dominant expert becomes more important; and under severe jitter or misalignment, the router suppresses risky cross-modal interaction and relies more on modality-specific evidence. The corresponding routing-log summaries and seed-level statistics are treated as supplementary evidence rather than as unverified claims in the main text.

\begin{figure}[t]
\centering
\safeincludegraphics[width=0.96\textwidth]{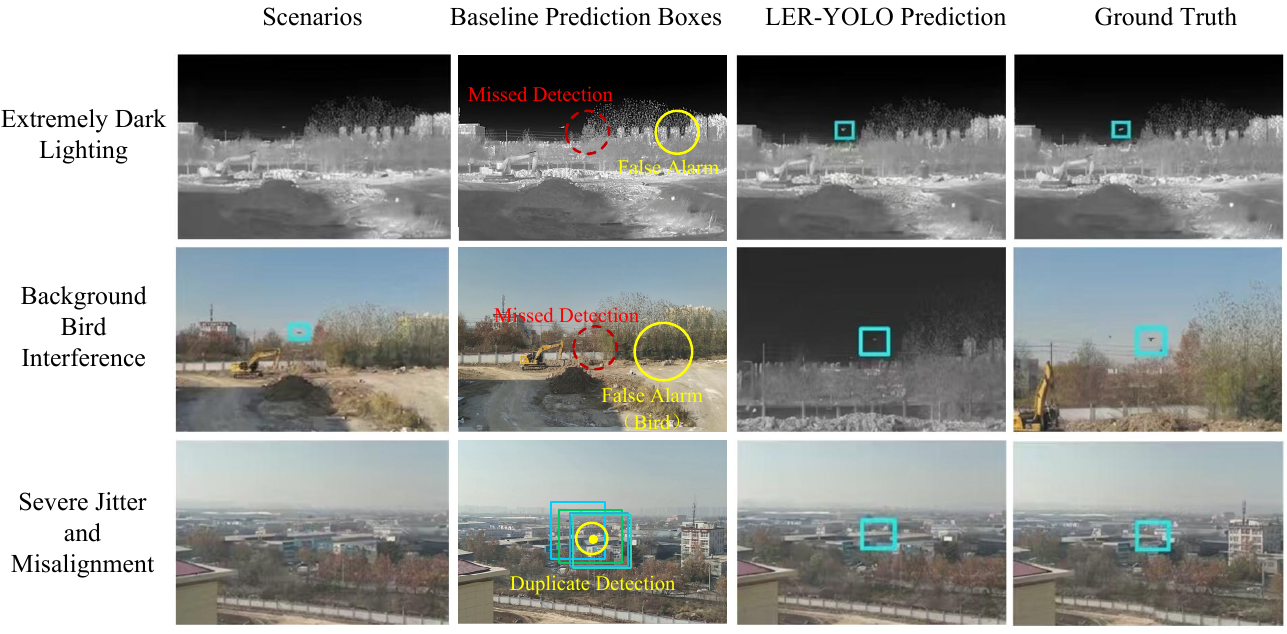}{Final Figure 5: qualitative detection comparison}
\caption{Qualitative detection comparison under challenging anti-UAV remote-sensing scenes. The baseline prediction column shows representative missed detection, false alarm, and duplicate detection cases, while the LER-YOLO prediction column shows more reliable localization under low illumination, bird-like clutter, and severe jitter or misalignment.\label{fig:qualitative}}
\end{figure}

The conceptual routing behavior in Figure~\ref{fig:routing} is complemented by qualitative detection examples in Figure~\ref{fig:qualitative}. These examples focus on visually challenging anti-UAV cases rather than on easy scenes. In these representative examples, the LER-YOLO prediction shows fewer missed detections, false alarms, and duplicate detections than the static-fusion baseline in low-illumination, bird-interference, and severe-jitter scenes. This qualitative evidence is consistent with the quantitative ablation and robustness analyses, which suggest that explicitly modeling alignment reliability can help reduce harmful fusion when cross-modal correspondence is unreliable.

\begin{figure}[t]
\centering
\safeincludegraphics[width=0.96\textwidth]{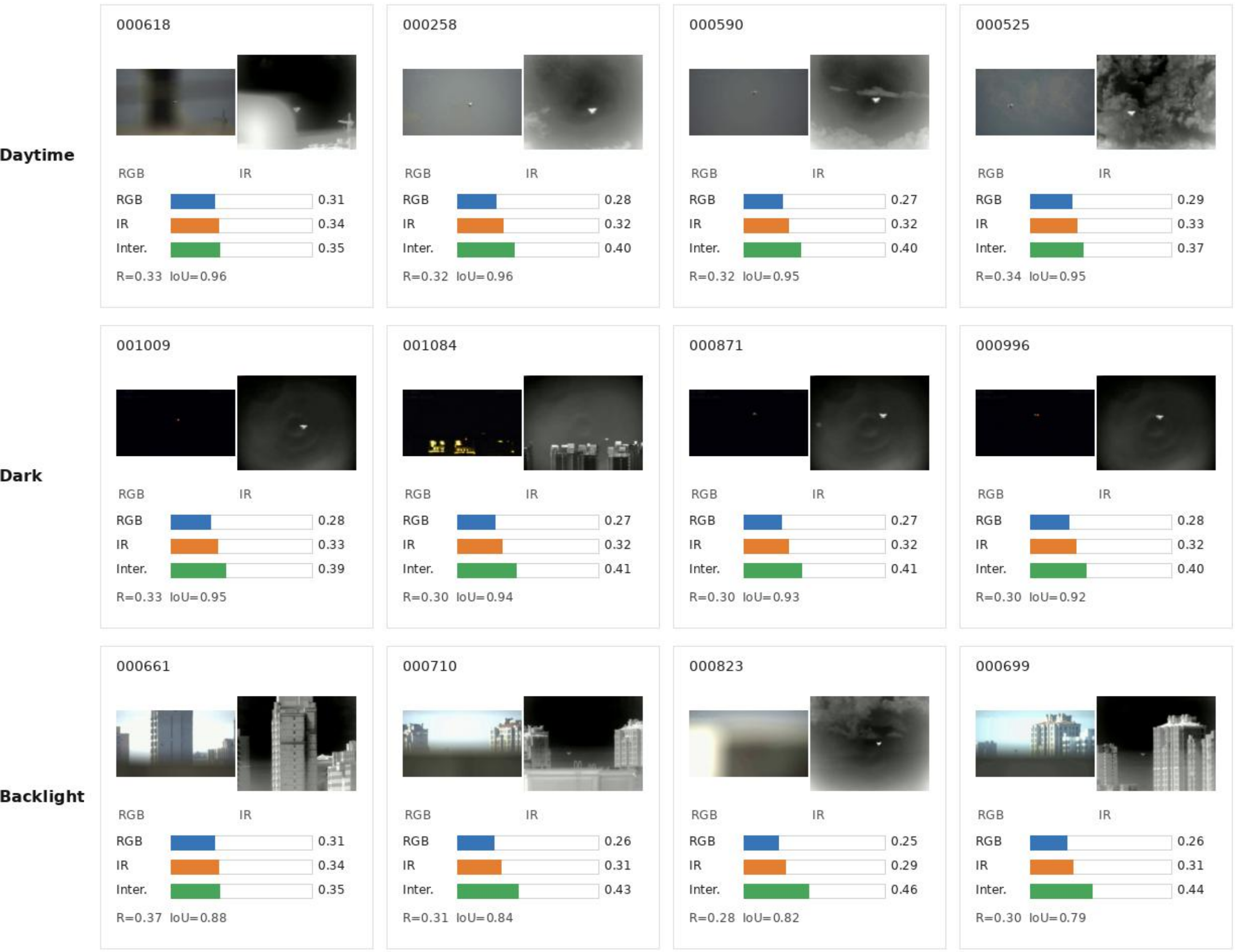}{Final Figure 6: representative scene-wise expert-weight examples}
\caption{Representative RGB-infrared examples and measured expert weights under daytime, dark, and strong-backlight scenes.\label{fig:scene_expert_weights}}
\end{figure}

To further examine real routing behavior, we summarize the actual scene-wise routing weights under daytime, dark, and strong-backlight scenes using the adopted $k=3$ setting. Table~\ref{tab:scene_expert_weights} reports the mean expert weights and target reliability scores for these three scene types, and Figure~\ref{fig:scene_expert_weights} shows representative RGB-infrared examples with the measured weights. The interaction expert receives the largest average weight in all three groups, while the infrared-dominant expert consistently obtains a slightly larger weight than the RGB-dominant expert. This trend is consistent with the role of infrared sensing under difficult illumination conditions, and the non-identical weights across scenes indicate adaptive rather than fixed routing.

\begin{table}[t]
\caption{Scene-wise actual routing weights of the Reliability-Guided Sparse MoE Fusion module under three representative scene types.}\label{tab:scene_expert_weights}
\centering
\footnotesize
\setlength{\tabcolsep}{4.0pt}
\renewcommand{\arraystretch}{1.12}
\begin{tabularx}{\textwidth}{@{}>{\raggedright\arraybackslash}p{0.18\textwidth}>{\centering\arraybackslash}p{0.08\textwidth}*{4}{>{\centering\arraybackslash}X}@{}}
\toprule
\textbf{Scene} & \textbf{$N$} & \textbf{$R_{target}$} & \textbf{$w_{rgb}$} & \textbf{$w_{ir}$} & \textbf{$w_{inter}$} \\
\midrule
Daytime & 1393 & 0.3272 & 0.2917 & 0.3301 & 0.3782 \\
Dark & 242 & 0.3053 & 0.2891 & 0.3319 & 0.3790 \\
Backlight & 39 & 0.3439 & 0.2884 & 0.3266 & 0.3850 \\
\bottomrule
\end{tabularx}
\vspace{2pt}
\begin{minipage}{\textwidth}
\footnotesize \textbf{Note:} $R_{target}$ is the average target-region reliability from the adopted $k=3$ model.
\end{minipage}
\end{table}

\subsection{Model Complexity}

Table~\ref{tab:complexity} reports model size and FLOPs under the reproduced profiling setting. The adopted final implementation uses top-$k$ routing with $k=3$ over the current three-expert pool; therefore, the present complexity is close to full-expert execution while retaining content-adaptive routing weights and an extensible top-$k$ formulation. Runtime metrics are not included in the main tables because the current implementation has routing overhead and speed is not used as an advantage claim.

\begin{table}[t]
\caption{Model complexity and detection performance under the reproduced profiling protocol.\label{tab:complexity}}
\centering
\footnotesize
\setlength{\tabcolsep}{2.8pt}
\renewcommand{\arraystretch}{1.12}
\begin{tabularx}{\textwidth}{@{}>{\raggedright\arraybackslash}X*{5}{>{\centering\arraybackslash}X}@{}}
\toprule
\textbf{Method} & \textbf{Params (M)} & \textbf{FLOPs (G)} & \textbf{P (\%)} & \textbf{R (\%)} & \textbf{\apfifty{} (\%)} \\
\midrule
YOLOv5s-RGB & 7.1 & 4.1 & $90.6\pm1.5$ & $76.1\pm1.6$ & $80.7\pm0.9$ \\
YOLOv5s-IR & 7.1 & 4.1 & $91.1\pm1.5$ & $76.5\pm1.6$ & $81.2\pm0.7$ \\
U-TA Static & 12.205 & 7.4 & $95.5\pm1.5$ & $83.6\pm1.3$ & $87.8\pm0.2$ \\
PM Static & 12.7 & 8.1 & $95.4\pm1.1$ & $83.0\pm1.0$ & $87.5\pm0.7$ \\
\ours{} & 12.21 & 7.5 & $96.9\pm0.5$ & $86.5\pm0.5$ & \textbf{$89.7\pm0.2$} \\
\bottomrule
\end{tabularx}
\vspace{2pt}
\begin{minipage}{\textwidth}
\footnotesize \textbf{Note:} FLOPs are measured under the reproduced profiling setting; U-TA Static and PM Static denote U-TA with static fusion and parameter-matched static fusion, respectively.
\end{minipage}
\end{table}

\subsection{Discussion and Limitations}

The experimental results support three observations. First, RGB-only and infrared-only detection both provide useful single-modality evidence on the \mbu{} benchmark, with infrared remaining slightly stronger under the infrared-reference protocol. This is consistent with UAV infrared remote-sensing studies, where thermal imagery remains useful under low illumination, cluttered backgrounds, and long-range observation. Second, naive fusion improves over single-modality detection, and alignment-aware plus reliability-aware fusion further improves the results. Third, the parameter-matched comparison shows that additional parameters alone do not explain the gain; routing decisions based on alignment reliability are important.

The method also has limitations. In the current three-expert implementation, the final default setting $k=3$ does not reduce execution to a strict sparse subset, so the computational advantage of sparse activation will become more meaningful after the expert pool is extended. The routing branch also introduces implementation overhead, and further kernel-level optimization may be required for embedded deployment. The synthetic-shift analysis is a controlled stress test rather than a complete replacement for real sensor-calibration changes. Moreover, the evaluation follows the public \mbu{} protocol, and broader validation on additional airborne, ground-to-air, and UAV-mounted RGB-infrared remote-sensing platforms would further test generalization. These limitations motivate future work on richer expert pools, lighter routing networks, hardware-aware sparse execution, and broader cross-platform validation.

\section{Conclusion}
\label{sec:conclusion}

This paper presents \ours{}, a reliability-aware sparse MoE fusion framework for misaligned RGB-infrared remote-sensing UAV target detection. The method connects Uncertainty-Aware Target Alignment (U-TA) with Reliability-Guided Sparse MoE Fusion, allowing the detector to decide when aligned RGB features should be trusted for cross-modal interaction and when modality-specific evidence should dominate. On the public \mbu{} benchmark, \ours{} achieves 89.7$\pm$0.2\% \apfifty{} across three independent seeds, with a best run of 89.9\%. Ablation, parameter-matched comparison, synthetic-shift testing, qualitative visualization, and model-complexity reporting suggest that the improvement is mainly associated with reliability-aware multimodal interaction rather than parameter growth alone. Future work will investigate richer expert pools, lighter routing networks, hardware-aware sparse execution, and broader validation on additional airborne, ground-to-air, and UAV-mounted RGB-infrared remote-sensing platforms.

\supplementary{The following supporting information is prepared for submission with the manuscript: Table S1, seed-level \apfifty{} results and training-log provenance; Table S2, synthetic-shift results by direction; Table S3, routing-weight or reliability-statistics summaries; and representative reliability-map or detection-case provenance used for the qualitative figures.}

\authorcontributions{Conceptualization, L.H.; methodology, L.H.; software, L.H.; validation, H.H.; formal analysis, L.H.; investigation, J.W.; resources, Y.H.; data curation, X.Z.; writing---original draft preparation, L.H.; writing---review and editing, Z.Y.; visualization, W.T.; supervision, Y.P.; project administration, X.Y.; funding acquisition, Y.P. All authors have read and agreed to the published version of the manuscript.}

\funding{This research was funded by the 2025 Equipment Comprehensive Research Project, grant number WJ2025C0401013, January 2026--December 2026; the 2025 Basic Frontier Innovation Project, grant number WJY202509, June 2025--May 2027; and the 2024 Second Batch Scientific Research Project entitled ``Research on Airborne-Vision-Based Counter Low-Slow-Small UAV Systems'', July 2024--June 2029.}

\institutionalreview{Not applicable. This study uses a public UAV image benchmark and does not involve human participants, animals, or personally identifiable private data.}

\informedconsent{Not applicable.}

\dataavailability{The data analyzed in this study are from the public Misaligned Bimodal UAV (MBU) benchmark introduced with MBUDet \cite{Ye2026MBUDet}. The MBU benchmark is derived from the public Anti-UAV RGB-infrared tracking dataset \cite{Jiang2021AntiUAV} through frame extraction and watermark removal. The MBUDet project and dataset information are available at \url{https://github.com/Yipzcc/MBUDet}. No new image dataset was created in this study. Supplementary files include seed-level summaries, routing-statistics summaries, and synthetic-shift evaluation records. Additional training logs and checkpoints are available from the corresponding author upon reasonable request, subject to repository and storage limitations.}

\acknowledgments{The authors thank the providers of the Anti-UAV and MBU benchmarks for making the RGB-infrared UAV data available for academic research. The authors also thank the Engineering University of PAP for research support.}

\conflictsofinterest{The authors declare no conflicts of interest.}

\abbreviations{The following abbreviations are used in this manuscript:\\
\noindent
\begin{tabular}{@{}ll}
AP & Average precision\\
IR & Infrared\\
LER & Local-reliability expert routing\\
MBU & Misaligned Bimodal UAV\\
MoE & Mixture of experts\\
Sparse MoE & Sparse mixture of experts\\
RGB & Red-green-blue visible imagery\\
RG & Reliability guidance\\
TA & Target alignment\\
UAV & Unmanned aerial vehicle\\
U-TA & Uncertainty-aware target alignment
\end{tabular}}

\begin{adjustwidth}{-\extralength}{0cm}
\reftitle{References}

\PublishersNote{}
\end{adjustwidth}

\end{document}